\documentclass[10pt,journal,compsoc]{IEEEtran}
\usepackage{amsmath,amssymb,amsfonts}
\usepackage{algorithmic}
\usepackage{graphicx}
\usepackage{textcomp}
\usepackage{framed,multirow}
\usepackage{latexsym}
\usepackage{amsmath}
\usepackage{threeparttable}
\usepackage{float}
\usepackage{array}
\usepackage{hhline}
\usepackage{colortbl}
\usepackage{stfloats}
\usepackage{booktabs}
\usepackage{multirow}
\usepackage{enumerate}
\usepackage{diagbox}
\usepackage{url}
\usepackage{hyperref}
\usepackage{makecell,rotating}
\usepackage{ulem}
\usepackage{subfigure}
\usepackage{graphicx}
\usepackage{makecell}
\usepackage{ulem}
\DeclareGraphicsExtensions{.pdf,.jpeg,.png,.jpg}

\graphicspath{{fig/},{fig/diagrams/},{fig/\biography/}}

\newcommand{\etal}{\textit{et al}.}

\definecolor{mygray}{gray}{.9}

\makeatletter

\newcommand{\Rmnum}[1]{\expandafter\@slowromancap\romannumeral #1@}
\makeatother

\ifCLASSOPTIONcompsoc
  \usepackage[nocompress]{cite}
\else
  \usepackage{cite}
\fi

\ifCLASSINFOpdf

\else

\fi

\begin{document}

\title{BroadCAM: Outcome-agnostic Class Activation Mapping for Small-scale Weakly Supervised Applications}
\author{Jiatai Lin, Guoqiang Han, Xuemiao Xu, Changhong Liang, Tien-Tsin Wong, C. L. Philip Chen, Zaiyi Liu, Chu Han

\IEEEcompsocitemizethanks{
\IEEEcompsocthanksitem Jiatai Lin, Guoqiang Han, Xuemiao Xu and C. L. Philip Chen are with the School of Computer Science and Engineering, South China University of Technology, Guangzhou, Guangdong, 510006, China.
\IEEEcompsocthanksitem Changhong Liang, Zaiyi Liu and Chu Han are with Guangdong Provincial Key Laboratory of Artificial Intelligence in Medical Image Analysis and Application, Guangzhou 510080, China and the Department of Radiology, Guangdong Provincial People's Hospital (Guangdong Academy of Medical Sciences), Southern Medical University, Guangzhou 510080, China.
\IEEEcompsocthanksitem Xuemiao Xu is with the School of Computer Science and Engineering at South China University of Technology, and also with State Key Laboratory of Subtropical Building Science, Ministry of Education Key Laboratory of Big Data and Intelligent Robot and Guangdong Provincial Key Lab of Computational Intelligence and Cyberspace Information.
\IEEEcompsocthanksitem Tien-Tsin Wong is with the Department of Computer Science and Engineering, The Chinese University of Hong Kong, Hong Kong, China.
\IEEEcompsocthanksitem Equal contribution: Jiatai Lin; Guoqiang Han.
\IEEEcompsocthanksitem Corresponding author: Zaiyi Liu; Chu Han.
}
}


\IEEEtitleabstractindextext{%
\begin{abstract}
Class activation mapping~(CAM), a visualization technique for interpreting deep learning models, is now commonly used for weakly supervised semantic segmentation~(WSSS) and object localization~(WSOL). It is the weighted aggregation of the feature maps by activating the high class-relevance ones. Current CAM methods achieve it relying on the training outcomes, such as predicted scores~(forward information), gradients~(backward information), etc. However, when with small-scale data, unstable training may lead to less effective model outcomes and generate unreliable weights, finally resulting in incorrect activation and noisy CAM seeds. In this paper, we propose an outcome-agnostic CAM approach, called BroadCAM, for small-scale weakly supervised applications. Since broad learning system (BLS) is independent to the model learning, BroadCAM can avoid the weights being affected by the unreliable model outcomes when with small-scale data. By evaluating BroadCAM on VOC2012 (natural images) and BCSS-WSSS (medical images) for WSSS and OpenImages30k for WSOL, BroadCAM demonstrates superior performance than existing CAM methods with small-scale data (less than 5\%) in different CNN architectures. It also achieves SOTA performance with large-scale training data. Extensive qualitative comparisons are conducted to demonstrate how BroadCAM activates the high class-relevance feature maps and generates reliable CAMs when with small-scale training data.
\end{abstract}

}

\maketitle

\IEEEdisplaynontitleabstractindextext
\IEEEpeerreviewmaketitle
\DeclareGraphicsExtensions{.pdf,.jpeg,.png,.jpg}
\graphicspath{{figs/},{figs/diagrams/}}

\IEEEraisesectionheading{\section{Introduction}\label{sec:intro}}
\IEEEPARstart{C}{lass} activation mapping (CAM)~\cite{CAM2016} has attracted great attention and emerged significant advancements in the past few years. It was first proposed to visualize the regions where the classification models focus for interpretability. Lately, it has been widely adopted for weakly supervised applications to reduce annotation efforts by using only image-level labels to achieve dense pixel-level predictions, including weakly supervised semantic segmentation (WSSS)~\cite{survey_WSSS_TPAMI2023} and weakly supervised object localization (WSOL)~\cite{survey_WSOL_TPAMI2022}.
\begin{figure}
	\centering
	\includegraphics[width=1\linewidth]{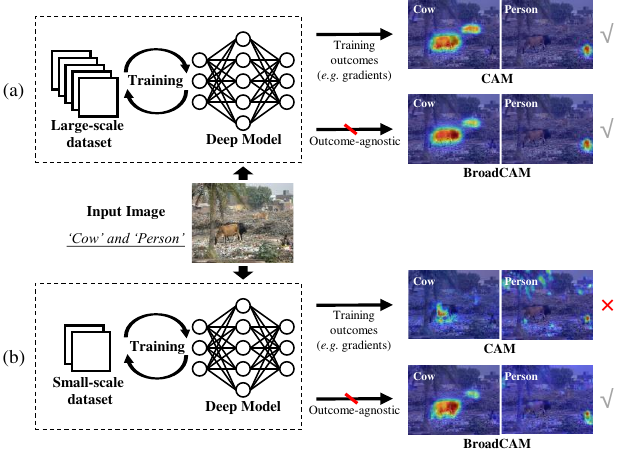}
	\caption{Comparison with CAM and our BroadCAM on large-scale data (a) and small-scale data (b). Existing CAM approaches rely on the training outcomes, and achieve good results on large-scale data but noisy results on small-scale data. The proposed BroadCAM is capable of generating stable class activation maps for the models trained on both large- and small-scale datasets.}\label{fig:idea}
\end{figure}

Most of the existing WSSS and WSOL approaches rely on the training outcomes to construct the correlation between the feature maps and their correlated categories, such as gradients for Grad-CAM~\cite{GradCAM2019} and prediction scores for Score-CAM~\cite{wang2020scoreCAM}, and generate weights for the feature map aggregation. Thanks to the large-scale open source data in the community (e.g., ImageNet~\cite{imagenet_cvpr09, ILSVRC15} and COCO~\cite{COCO}), current CAM-based weakly supervised applications have achieved outstanding performance as shown in Fig.~\ref{fig:idea}~(a). However, when with small-scale data, the model training may become unstable, which leads to unreliable training outcomes and finally generates noisy weights. That is the reason why most of the current CAM approaches that depend on the training outcomes fail to generate reliable CAM seeds for weakly supervised applications, shown in Fig.~\ref{fig:idea}~(b). And this situation cannot be underestimated. Because in the deep learning era, a tailored dataset for each specific task or scenario is indispensable, but most of the large-scale datasets were collected for the general natural scenes. Collecting massive well-labeled data from scratch is extremely difficult and labor-intensive. Therefore, a reliable approach for small-scale weakly supervised applications is crucial for reducing the cost and efforts of dense annotations and accelerating the annotation process.

Since the unreliable training outcomes are the reason why existing CAM approaches fail to generate appropriate CAM seeds with small-scale data. The most intuitive idea is to make the CAM generation process independent to the training outcomes. So the key problem to be solved is to re-construct the correlation between the feature maps and their correlated categories in an alternative way while ensuring the high and low class-relevance feature maps are activated and deactivated, respectively. To this end, we propose an outcome-agnostic CAM approach, called BroadCAM, for small-scale weakly supervised applications. Broad learning system (BLS)~\cite{BLS2018} is introduced as an independent classifier to re-construct the correlation. Thanks to the outcome-agnostic nature, BLS can avoid the weights being affected by the unreliable model outcomes while successfully activating the high class-relevance feature maps and deactivating the low class-relevance ones. Thanks to the robustness of BLS in handling small-scale data~\cite{PDBL}, the training of BLS is stable and the generated weights are more reliable when with insufficient training samples. To further bridge the gap between image-level labels and dense pixel-level predictions, we introduce more information by aggregating multi-layer feature maps. To the best of our knowledge, we are the first CAM approach to handle small-scale weakly supervised applications.

We conduct extensive experiments to evaluate the quality of CAM seeds for WSSS and WSOL on three datasets, including two natural image ones and a medical image one. Quantitative results demonstrate that BroadCAM greatly outperforms the most representative CAM methods across different model architectures and training strategies in small-scale data (less than 5\%). In the entire data gamut experiments, the CAM seed generated by BroadCAM also consistently achieves state-of-the-art (SOTA) performance for both WSSS and WSOL on all the datasets. By qualitatively comparing with SOTA CAM approaches, BroadCAM is highly related to the corresponding categories with more complete activation and fewer noises when with small-scale training data. In the meanwhile, BroadCAM also shows superior activation results when with large-scale data. Furthermore, by visualizing the relationship between the weights and feature maps, we observe that BroadCAM activates more high class-relevance feature maps and fewer low class-relevance ones than conventional CAM approaches in small-scale data. To summarize, BroadCAM is an effective and flexible CAM approach for weakly supervised applications which is less susceptible to the dataset size. The contributions of this paper are summarized as follows.

\begin{itemize}
  \item This paper is the first study that focuses on small-scale weakly supervised applications. We provide a feasible solution by making the CAM generation process independent to training outcomes when they are not reliable.
	\item We introduce a novel outcome-agnostic CAM named BroadCAM for small-scale WSSS and WSOL. BLS guarantees the reliability of the weights in small-scale data. Multi-layer feature map aggregation bridges the gap between weak supervision and dense prediction.
	\item Data gamut experiments on both WSSS and WSOL demonstrate the superiority and robustness of BroadCAM compared with the most representative CAM approaches. BroadCAM achieves SOTA quantitative performance in two natural image datasets and one medical image dataset.
	\item Qualitative comparisons demonstrate the robustness of BroadCAM on all datasets with small-scale data. Visualization of the relationship between CAM weights and feature maps shows the reliability of CAM weights generated by BroadCAM.
\end{itemize} 

\section{Related Works}\label{sec:related}
\subsection{CAM-based Weakly Supervised Applications}\label{sec:related_ws}
Weakly supervised applications, including WSSS and WSOL, leverage image-level labels to achieve dense pixel-level predictions~\cite{survey_WSSS_TPAMI2023, survey_WSOL_TPAMI2022}. The major challenge is the huge information gap between weak supervision and dense prediction. Class activation mapping (CAM) technique~\cite{CAM2016} provides a feasible solution to overcome this challenge. Since the feature maps of the classification model are able to reveal the position and semantically related areas of the target classes. Therefore, CAM-based approaches have become the mainstream for weakly supervised applications.

Current WSSS and WSOL approaches apply various training strategies or introduce more useful information to push the classification task toward the segmentation/localization tasks. To avoid the classification model only focusing on the most distinguishable regions, the dropout strategy~\cite{Choe2019ADL,WSSS2022han} is introduced to deactivate the most discriminative areas, forcing the model to learn from non-predominant regions and generate more complete CAM seeds. To solve the incorrect activations in different scales, SEAM~\cite{SEAM_2020_CVPR} is proposed to reconstruct the consistency over different rescaling transformations in a self-supervised manner. To encourage finer local details, PuzzleCAM~\cite{Puzzle2021} and L2G~\cite{L2G_2022_CVPR} are proposed to train the classification model jointly by the whole image and the cropped/tessellated images. To introduce additional object information, saliency-guided methods have been proposed to exploit the saliency maps as background cues~\cite{fan2020cian, Lee2019FickleNetWA}, pseudo-pixel supervision~\cite{EPS2021} and class-specific attention cues~\cite{SGAN2020}. Moreover, existing studies also introduce information from other aspects to bridge the information gap, such as contrastive self-attention~\cite{ACoL2018}, data augmentation~\cite{yun2019cutmix} and domain adaptation~\cite{zhu2022weakly}.

By reviewing CAM-based weakly supervised techniques, we find that current approaches are basically constructed on top of the deep models trained on large-scale training data. However, pursuing sufficient high-quality labeled data is labor-intensive, even for image-level labels. Although the foundation models are the trend of future artificial intelligence (AI)~\cite{deeplearningsurvey2021}, the paradigm of current AI models remains one dataset for one specific task. Small-scale data will still be an inevitable problem for a long time in the deep learning era. Gao~\etal~\cite{LUSS2022} presents a pioneering study on large-scale unsupervised semantic segmentation (LUSS) with a released benchmark. However, even though they have achieved SOTA performance on LUSS, the precision of the results is still insufficient in practice due to the lack of supervision. In this paper, we first raise the importance of small-scale weakly supervised applications and propose a novel BroadCAM for them.

\subsection{Class Activation Mapping~(CAM)}
\label{sec:related_cam}
\begin{figure}
	\centering
	\includegraphics[width=.975\linewidth]{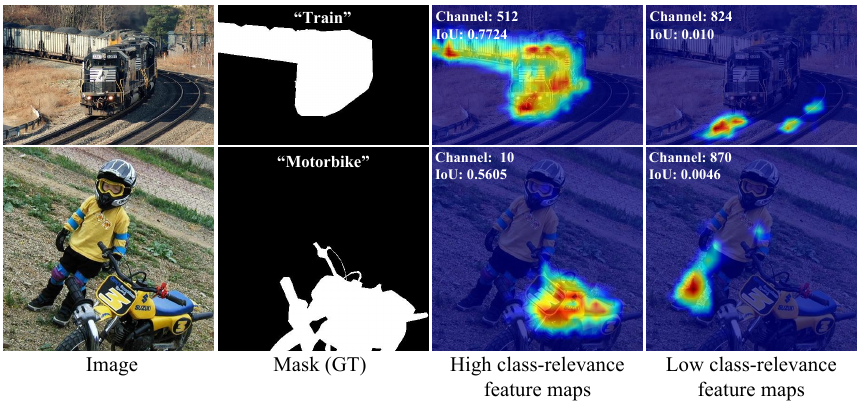}
	\caption{Examples of high and low class-relevance feature maps. For each feature map, we calculate the IoU between the activated region~(threshold: 0.1) and the ground truth mask to measure the relevance between each feature map and the corresponding category.}
	\label{fig:featuremaps}
\end{figure}

Besides the training strategy, CAM techniques are also crucial for WSSS and WSOL. Essentially, CAM seed generation is a weighted aggregation of the feature maps. The quality of CAM seeds strongly relies on the robustness of the weights, which determine whether higher class-relevance feature maps (shown in Fig.~\ref{fig:featuremaps}) can contribute more activation. Current CAM methods generate weights by constructing the correlation between the feature maps and their corresponding categories based on the training outcomes, which can be categorized into two types: gradient-based and gradient-free methods.

\subsubsection{Gradient-based CAM}
GradCAM~\cite{GradCAM2019} is the first CAM approach that leverages gradient information to improve the flexibility of CAM, which has been widely applied to WSSS and WSOL. Later, various gradient-based methods have been proposed to improve the granularity by the smoothing technique, such as GradCAM++~\cite{GradCAM++2018}, smooth GradCAM++~\cite{omeiza2019smooth} and etc. To make the CAM seeds more complete for weakly supervised applications, LayerCAM~\cite{LayerCAM2021} is proposed to introduce additional coarse-to-fine information by weighted aggregating multi-layer feature maps. BagCAM~\cite{BagCAMs} identifies globally discriminative areas by deriving a set of regional localizers from the well-trained classifier, rather than solely activating regional object cues.

\subsubsection{Gradient-free CAM}
To avoid the saturation and false confidence problems of gradient, Score-CAM~\cite{wang2020scoreCAM} utilizes forward information~(predicted score) to replace backward information (gradient) to achieve less noisy CAM seeds.
Lately, several score-based CAM methods have been introduced to advance Score-CAM to generate more complete visual explanations of deep models by the smooth operation~(SS-CAM~\cite{SSCAM2020}) and integration operation~(IS-CAM~\cite{ISCAM2020}). To produce high-quality and class-discriminative localization maps, Ablation-CAM~\cite{AblationCAM2020} employs ablation analysis to measure the significance of individual feature map units to obtain the weights for CAM generation. Relevance-CAM~\cite{Relevance-CAM2021} overcomes the shattered gradient problem at shallow layers by utilizing a relevance propagation process.

Although current CAM techniques bridge the gap between weak supervision and dense predictions, they are still based on an underlying assumption of the presence of large-scale data. When with small-scale data, the performance of current training outcome-based CAM methods may suffer from a dramatic decrease since unstable training will lead to unreliable training outcomes. To overcome such difficulty, we propose an outcome-agnostic CAM approach for small-scale data by breaking the dependency on training outcomes when generating CAM seeds.

\subsection{Broad Learning System~(BLS)}
\label{sec:related_bls}
Broad learning system~(BLS)~\cite{BLS2017, BLS2018, review_BLS_2022} is a flat and lightweight classification network without the need for deep architecture. It constructs the correlation fleetly between the image/features and labels by feature mapping and enhanced mapping to avoid the retraining process. In the last few years, various modified BLS have been developed to improve the broad learning architecture and the feature mapping manner, such as Fuzzy-BLS~\cite{FuzzyBLS}, TDFW-BLS~\cite{TDFW-BLS2019}, stacked-BLS~\cite{stackedBLS2021}, Recurrent-BLS~\cite{RecurrentBLS2020} and etc. BLS is also widely applied in various applications such as hyperspectral image classification~\cite{BLS_application_2022} and micro-robotic control~\cite{xu2022MC}.

In our previous studies~\cite{PDBL,deng2023feddbl}, we also explore the capability of BLS on small-scale classification. Our proposed pyramidal deep-broad learning (PDBL) demonstrates more stable and superior performance in different CNN architectures compared with the conventional deep learning manner on small-scale data. Even if the model training is unstable with small-scale data, BLS can still successfully construct the correlation between the features and the labels. Therefore, we further extend BLS to the problem of small-scale CAM generation and finally empower small-scale WSSS and WSOL. 
\begin{figure*}~\centering
	\includegraphics[width=1\linewidth]{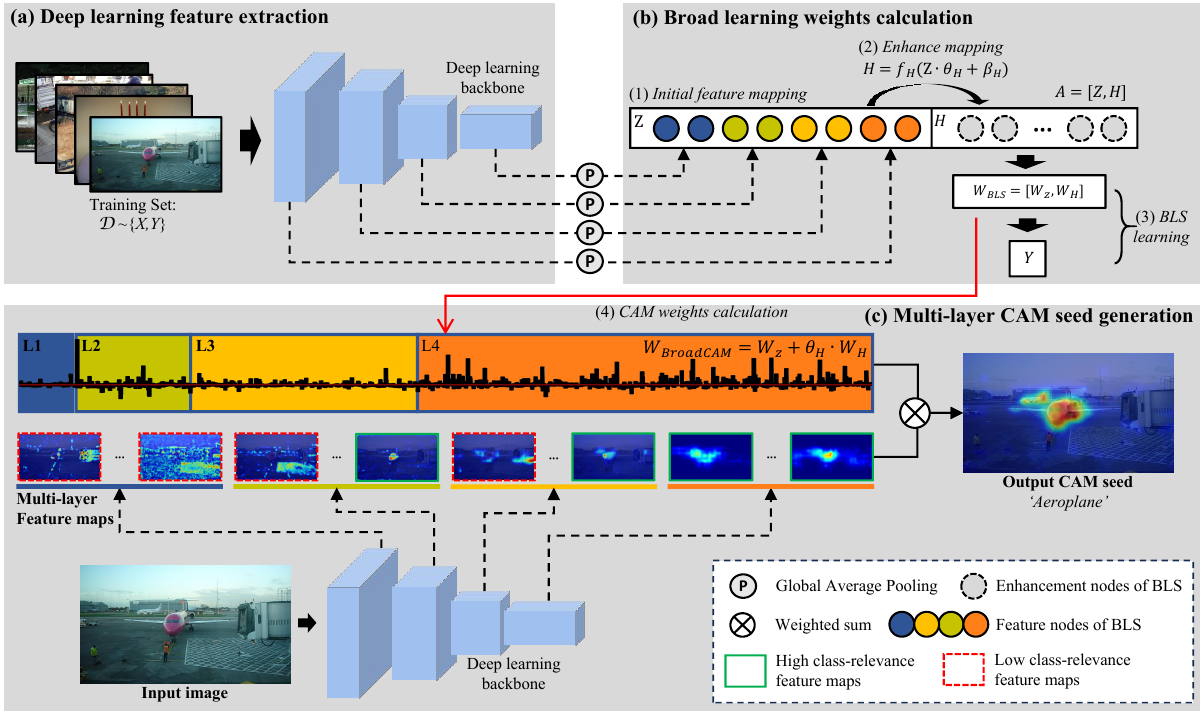}
	\caption{The model framework of our proposed BroadCAM, which includes three steps. (a) Deep learning feature extraction: We first extract the features of multiple layers from a deep learning model. (b) Broad learning weights calculation: A broad learning system with only one-round learning is applied to generate the BroadCAM weights $W_{BroadCAM}$. (c) Multi-layer CAM seed generation: CAM seed is generated by a weighted aggregation of the multi-layer feature maps. We define the high class-relevance feature map as the one with a larger overlap with the ground truth label.}
	\label{fig:workflow}
\end{figure*}

\section{BroadCAM}\label{sec:3.1}
In this paper, we propose an outcome-agnostic CAM approach for small-scale WSSS and WSOL, called BroadCAM. Since CAM seed generation is essentially a weighted aggregation of the feature maps. Generating high-quality CAM seeds depends on whether CAM weights can activate the higher class-relevance feature maps and deactivate the lower ones.
For small-scale weakly supervised applications, outcome-based CAM approaches generally fail to generate reliable weights due to unstable training outcomes. Thus, we design a novel outcome-agnostic CAM approach by making the weights generation process independent of the training outcomes. Fig.~\ref{fig:workflow} demonstrates the workflow of the proposed BroadCAM. And the details of BroadCAM are shown as follows.

\subsection{Deep Learning Feature Extraction}\label{sec:3.1.1}
Given a training set~$\mathcal{D}\sim\{{X},{Y}\}$, where ${X}$ and ${Y}$ are the training samples and the corresponding image-level labels, we first train a classification model~$f_{DL}$ to learn domain-specific knowledge from the dataset, just like the other CAM approaches. Then we use the classification model as the deep feature extractor. To further bridge the gap and introduce more information, we extract deep features from multiple layers.
\begin{equation}\label{eq:cam}
	\mathcal{F}_{1}, \mathcal{F}_{2}, ..., \mathcal{F}_{j} = f_{DL}(x, \theta_{DL})
\end{equation}
where $j$ is the index of the layers. $x\in {X}$, $\theta_{DL}$ and $\mathcal{F}_{j}$ represent a sample $x$ in the training set~${X}$, parameters of deep learning model and the feature maps of the $j^{th}$ layer. Since the deep learning architecture in BroadCAM only serves for feature extraction. Any training strategy to enhance the feature maps can be applied to this framework. And we will discuss this in the experiments.

\subsection{Broad Learning Weights Calculation}\label{sec:3.1.2}
Next, we construct the correlation between the deep features and the corresponding categories and generate the weights for each feature map. Inspired by our previous studies~\cite{PDBL,deng2023feddbl}, a broad learning system (BLS) is a lightweight and flat architecture that is a robust classifier for small-scale data. Thus, we introduce BLS to generate CAM weights for CAM seed generation. In the BLS, we first map the deep feature maps to broad features to fit the broad learning architecture by initial feature mapping. Then, we establish the enhancement nodes to expand the feature broadly by enhanced mapping. And we solve the optimization problem of BLS with extended broad learning feature nodes by ridge regression algorithm. Finally, we can transform the BLS parameters to obtain the weights for CAM seed generation.

\textbf{(1) Initial Feature Mapping}: For the feature maps $\mathcal{F}_{j}$ extracted from the $j^{th}$ layer, we first apply global average pooling~(GAP)~$f_{GAP}$ to squeeze the deep learning features by:
\begin{equation}\label{eq:GAP}
	\begin{split}
		z_{j} = f_{GAP}(\mathcal{F}_{j})
	\end{split}
\end{equation}
where $z_{j}$ represents the squeezed features of the $j^{th}$ layer. Then we concatenate the multi-layer features of each sample into a broad feature vector~$\mathbf{z}$ as the feature node to fit the BLS architecture:
\begin{equation}
	\mathbf{z} = [z_{1}, z_{2}, ..., z_{j}]^T.
\end{equation}

We form a broad feature matrix ${Z}$ of all the training samples by:
\begin{equation}
	{Z} = [\mathbf{z}^{1}, \mathbf{z}^{2}, ..., \mathbf{z}^{i}]^T
\end{equation}
where $i$ is the index of samples.

\textbf{(2) Enhance Mapping}: Next, we perform an enhanced mapping on the feature nodes to expand the broad structure widely and improve the BSL optimization.
\begin{equation}\label{eq:enhance_mapping}
	\begin{split}
		{H} = f_{{H}}({Z}\times \theta_{{H}}+\beta_{{H}}).
	\end{split}
\end{equation}
where $f_{{H}}$ is a linear activation function. $\theta_{{H}}$ and $\beta_{{H}}$ represent the weights and bias of enhance mapping.
$\theta_{{H}}$ is used to group the feature nodes with similar importance. Different from the original BLS design with randomly initialized weights $\theta_{{H}}$, $\theta_{{H}}$ is now initialized by the class-relevance of the features ${Z}$. To acquire the class relevance, we first solve BLS using the initial feature nodes to calculate the parameters $W_{init}$ by:
\begin{equation}\label{eq:first_ridge}
	W_{init}=(\lambda I+{Z}{Z}^T)^{-1} {Z}^{T} Y
\end{equation}
where $I$ and $\lambda$ denote the unit matrix and the hyper-parameter of the L2 norm constraint for $W_{init}$. Then we use $W_{init}$ to initialize $\theta_{{H}}$.

\textbf{(3) BLS Learning}: Then we utilize all the broad features to learn BLS to construct the relationship between the features and the corresponding image-level labels~$Y$. We first concatenate the feature nodes and enhancement nodes together into an expanded broad feature matrix, denoted as ${A} = [{Z},{H}]$. Next, we employ the ridge regression algorithm to calculate the parameters of BLS as follows:
\begin{equation}
	\begin{split}
		W_{BLS} = (\lambda I+{A}{A}^T)^{-1} {A}^{T} Y
	\end{split}
\end{equation}
where $W_{BLS}$ is the BLS parameters which can be split into two parts $W_{{Z}}$ and $W_{{H}}$ for the feature nodes and enhancement nodes, respectively.

\textbf{(4) CAM Weights Calculation}: To use the BLS parameters to generate the CAM weights for each feature map, we transform the formulation of BLS as follows:
\begin{equation}
	\begin{split}
		Y&={A}\cdot W_{BLS} =[{Z},{H}]\times [W_{{Z}}, W_{{H}}]^T\\
		&={Z}\times W_{{Z}}+{H}\times W_{{H}}\\
		&={Z}\times W_{{Z}}+({Z}\times \theta_{{H}}+\beta_{{H}})\times W_{{H}}\\
		&={Z}\times (W_{{Z}}+\theta_{{H}}\times W_{{H}}) + \beta_{{H}}\times W_{{H}}\\
		denote:&={Z}\times W_{BroadCAM} + \sigma
	\end{split}\label{Eq:channel_weight}
\end{equation}
where $\sigma=\beta^{{H}}\times W_{{H}}$ is the bias of the BLS and $W_{BroadCAM}=W_{{Z}}+\theta_{{H}}\times W_{{H}}$ represents the weights that directly connect the feature nodes to the predicted classification results, quantifying the relevance between deep feature maps and categories. Since each value of the broad feature vector ${Z}$ is calculated by GAP from the corresponding deep feature map of ${F}$. The CAM weights of ${Z}$ are equivalent to the CAM weights of ${F}$. It is the channel-wise weights utilized for aggregating feature maps to generate class activation maps.

\subsection{Multi-layer CAM Seeds Generation}\label{sec:3.1.3}
Finally, we can use $W_{BroadCAM}$ to generate CAM seeds in the weights aggregation manner by:
\begin{equation}\label{eq:linear_combination}
	M_{BroadCAM} = ReLU([\mathcal{F}_1, \mathcal{F}_2, ..., \mathcal{F}_i] \times W_{BroadCAM})
\end{equation}
where $ReLU$ is the non-linear activation function. The proposed BroadCAM can aggregate the feature maps from multiple layers. 

\section{Datasets}\label{sec4}
In this paper, we evaluate the proposed BroadCAM on three datasets including the VOC2012 dataset~(WSSS), BCSS~(WSSS), and OpenImages30k~(WSOL), which are detailed in this section. The distributions of three datasets are demonstrated in Table~\ref{tab:dataset_distribution}.

\subsection{VOC2012 Dataset}
VOC2012 dataset\footnote{\url{http://host.robots.ox.ac.uk/pascal/VOC/voc2012/}} is the standard dataset provided by The PASCAL Visual Object Classes Challenge~\cite{VOC2012}, which has been used for weakly supervised semantic segmentation~(WSSS) tasks in recent years. VOC2012 dataset comprises $20$ foreground categories, encompassing Car, Bus, Bicycle, etc. It can be split into training set~(1464 images), validation set~(1449 images) and test set~(1456 images). Following the previous WSSS study~\cite{Puzzle2021}, the training set has been extended to an augmented training set which includes $10,582$ training samples. Image-level labels are provided for model training and pixel-level annotations are for evaluation. In our experiments, we use the augmented training set with different proportions to train the models and evaluate the CAM seed in the original training set and validation set.

\begin{table}[t]
\centering
\caption{Data distribution of VOC2012, BCSS-WSSS and OpenImages30k datasets.}
\begin{tabular}{c|c|c|c|c|c}
  \hline
  \multirow{2}{*}{Task} & \multirow{2}{*}{Dataset} & \multirow{2}{*}{\makecell{Num. of\\categories}} & \multicolumn{3}{c}{Num. of samples}    \\ \cline{4-6}
  & & & train & val & test \\ \hline
  \multirow{2}{*}{WSSS}
  & VOC2012~\cite{VOC2012} & 20 & \makecell{10,582\\ (aug)} & 1,449 & 1,456\\ \cline{2-6}
  & BCSS-WSSS~\cite{WSSS2022han} & 4 & 23,422 & 3,418 & 4,986 \\ \hline
  \multirow{1}{*}{WSOL}
  & \makecell{OpenImages30k\\\cite{choe2020cvpr, choe2022tpami}} & 100    & 29,819  & 2,500  & 5,000 \\
  \hline
\end{tabular}\label{tab:dataset_distribution}
\end{table}

\subsection{BCSS-WSSS Dataset}
Breast Cancer Semantic Segmentation~(BCSS) dataset~\cite{BCSS_2022}\footnote{\url{https://bcsegmentation.grand-challenge.org/}} is a well-labeled dataset for histopathological tissue semantic segmentation task of breast cancer. It consists of $151$ representative regions of interest (ROIs) which were selected from $151$ whole slide images~(WSIs) and labeled by pathologists, pathology residents, and medical students. In this dataset, pixel-level annotations were provided for fully supervised model training, including $4$ predominant classes~(e.g.,~tumor, stroma, lymphocyte-rich regions and necrosis.), $6$ non-predominant classes~(e.g.,~artifacts, blood, etc.) and $8$ challenging classes~(e.g.~plasma cells, lymph vessels, etc.)

Based on the BCSS dataset, our previous study~\cite{WSSS2022han} recreates a new dataset~(BCSS-WSSS dataset) for weakly supervised semantic segmentation, which cropped the patches from original ROIs and obtained the corresponding image-level labels from the pixel-level annotations. In the BCSS-WSSS dataset, $4$ foreground categories were defined based on the predominant classes of the BCSS dataset, including Tumor (TUM), Stroma (STR), Lymphocytic infiltrate (LYM), Necrosis (NEC).

\subsection{OpenImages30k Dataset}
OpenImages30k dataset~\cite{choe2020cvpr, choe2022tpami}\footnote{\url{https://github.com/clovaai/wsolevaluation}} is curated for weakly supervised object localization (WSOL) task based on the OpenImagesV5 dataset~\cite{OpenImagesV5}\footnote{\url{https://storage.googleapis.com/openimages/web/download_v5.html}}. It consists $100$ categories with balanced number of samples in each category. Samples of OpenImages30k were selected randomly from the OpenImagesV5 dataset which can be split into training set~(29819 images), validation set~(2500 images) and test set~(5000 images), respectively. In this dataset, the image-level labels are provided in the training set for classification model training with different proportions and pixel-level annotations were created for evaluation.

\begin{table*}[t]
\centering
\caption{Comparison of our proposed BroadCAM and existing CAM approaches on VOC2012 and BCSS-WSSS datasets with different proportions of training data. mIoU and FwIoU are used for the evaluation of VOC2012 and BCSS-WSSS, respectively. \textbf{Bold} represents the best performance.}
\begin{tabular}{c|c|c|c|c|c|c|c|c|c|c|c|c}
  \hline
   \multirow{2}{*}{Dataset}& \multicolumn{1}{c|}{\multirow{2}{*}{\makecell[c]{Training strategy\\(Architecture)}}} & \multicolumn{1}{c|}{\multirow{2}{*}{Split}} & \multirow{2}{*}{CAM method} &\multicolumn{9}{c}{Performance of CAM seeds with different proportions of training samples}\\ \cline{5-13}
   && &  & 1\%   &2\%   &5\%   &8\%   &10\%
   & 20\%   &50\%   &80\%   &100\% \\ \hline
   \multirow{16}{*}{\makecell{VOC2012\\(natural image)}}
   & \multirow{8}{*}{\makecell[c]{Puzzle~\cite{Puzzle2021}\\(ResNeSt101)}}& \multirow{4}{*}{train} & CAM
   & 34.13   & 37.18   & 46.51   & 47.44   & 49.75   & 53.55   &  \textbf{58.15}  & \textbf{58.89}   & \textbf{59.55}   \\
   & &   & GradCAM
   & 33.89   & 36.71   & 46.05   & 47.05   & 49.40   & 53.10   & 57.82   & 58.85   & 59.53    \\
   & &   & LayerCAM
   & 41.18   & 43.46   & 46.27   & 47.03   & 49.54   & 45.61   & 51.18   & 54.12   & 54.64    \\
   & &   & BroadCAM
   & \textbf{46.45}   & \textbf{47.02}   & \textbf{48.57}   & \textbf{49.55}   & \textbf{51.51}   & \textbf{54.22}   & 55.90   & 58.39   & 58.52   \\ \cline{3-13}

   & & \multirow{4}{*}{val}  & CAM
   & 33.85   & 36.83   & 46.28   & 47.00   & 49.41   & 53.58   &  \textbf{57.33}  & \textbf{58.99}   & \textbf{59.00}    \\
      & &   & GradCAM
   & 33.73   & 36.30   & 45.89   & 46.77   & 49.04   & 52.72   & 57.00   & 58.96   & 58.90    \\
   & &   & LayerCAM
   & 41.47   & 43.22   & 45.75   & 46.23   & 48.61   & 45.67   & 51.38   & 53.90   & 54.43    \\
   & &   & BroadCAM
   & \textbf{46.55}   & \textbf{46.78}   & \textbf{48.54}   & \textbf{49.18}   & \textbf{51.23}   & \textbf{53.78}   & {54.72}   & 57.33   & 57.41   \\ \cline{2-13}

      & \multirow{8}{*}{\makecell[c]{Puzzle~\cite{Puzzle2021}\\(ResNeSt269)}}& \multirow{4}{*}{train} & CAM
   & 30.11   & 38.59   & 45.25   & 51.55   & 51.17   & 54.26   & 56.66   & \textbf{58.89}   & \textbf{58.85}    \\
   & &   & GradCAM
   & 29.97   & 37.92   & 44.58   & 50.62   & 50.04   & 53.93   & 56.02   & 58.53   & 58.47    \\
   & &   & LayerCAM
   & 41.81   & 44.52   & 45.88   & 50.60   & 48.85   & 53.12   & 52.89   & 55.01   & 55.47    \\
   & &   & BroadCAM
   & \textbf{49.37}   & \textbf{49.00}   & \textbf{49.78}   & \textbf{54.38}   & \textbf{52.26}   & \textbf{57.03}   & \textbf{56.85}   & 58.69   & 57.89   \\ \cline{3-13}

   & & \multirow{4}{*}{val}  & CAM
   & 29.95   & 38.96   & 44.85   & 51.53   & 51.56   & 54.27   & 56.11   & \textbf{58.39}   & \textbf{58.45}    \\
      & &   & GradCAM
   & 29.75   & 38.57   & 44.36   & 50.77   & 50.56   & 53.81   & 55.56   & 58.00   & 58.03    \\
   & &   & LayerCAM
   & 41.41   & 44.06   & 45.82   & 50.60   & 49.20   & 52.12   & 52.02   & 54.52   & 54.74    \\
   & &   & BroadCAM
   & \textbf{49.87}   & \textbf{48.31}   & \textbf{49.63}   & \textbf{54.06}   & \textbf{52.17}   & \textbf{56.20}   & \textbf{56.67}   & 57.96   & 57.49   \\ \hline
   \multirow{8}{*}{\makecell{BCSS-WSSS\\(medical image)}}
   & \multirow{8}{*}{\makecell[c]{PDA\cite{WSSS2022han}\\(ResNet38)}}& \multirow{4}{*}{val} & CAM
   & 48.38   & 60.17   & 65.91   & 68.24   & 69.18   & 69.55   & 69.69   & 70.11  & 70.64   \\
   & &   & GradCAM
   & 48.37   & 59.81   & 65.96   & 68.30   & \textbf{69.28}   & 69.87   & 69.85   & 70.32  & 70.79    \\
   & &   & LayerCAM
   & 52.93   & 64.46   & 65.92   & 66.86   & 67.35   & 68.90   & 69.51   & 69.04  & 68.78    \\
   & &   & {BroadCAM}
   & \textbf{59.92}   & \textbf{64.77}   & \textbf{66.86}   & \textbf{69.01}   & 68.11   & \textbf{71.73}   & \textbf{71.03}   & \textbf{71.08}   & \textbf{71.54}   \\ \cline{3-13}
   &&\multirow{4}{*}{test} & CAM
   & 55.91   & 67.53   & 71.80   & 73.71   & 73.44   & 73.04   & 73.18   & 73.93  & 74.14   \\
   & &   & GradCAM
   & 55.84   & 67.43   & \textbf{72.09}   & 73.98   & \textbf{73.69}   & 73.38   & 73.37   & 74.13  & 74.26    \\
   & &   & LayerCAM
   & 59.09    & 67.64   & 68.90   & 69.78   & 70.08   & 71.11   & 71.72   & 71.46  & 70.84    \\
   & &   & {BroadCAM}
   & \textbf{66.63}   & \textbf{70.50}   &  71.79  & \textbf{74.09}   & 72.41   & \textbf{74.59}   & \textbf{74.14}   & \textbf{74.47}   & \textbf{74.93}   \\ \cline{3-13}
  \hline
\end{tabular}\label{tab:quantitative_WSSS}
\end{table*} 
\section{Experiments}\label{sec:exp}
In this section, we first show the evaluation metrics in our experiments in Section~\ref{Sec:metrics}. Then, we evaluate the quality of the CAM seeds generated by various CAM techniques on both WSSS and WSOL in three datasets in Section~\ref{sec:exp_WSSS} and Section~\ref{sec:exp_WSOL}. Next, we comprehensively analyze the reliability of the CAM weights in Section~\ref{Sec:effectiveness_analysis}. Finally, we conduct an ablation study on the multi-layer design of BroadCAM in Section~\ref{Sec:ablation}.

\subsection{Evaluation Metrics}\label{Sec:metrics}
In this paper, we conduct the comparison of CAM methods on VOC2012 dataset~\cite{VOC2012}~(WSSS, natural image), BCSS-WSSS~\cite{WSSS2022han}~(WSSS, medical image) and OpenImages30k~\cite{choe2020cvpr, choe2022tpami}~(WSOL, natural image), respectively. We employ three metrics to evaluate the performance of CAM seeds:
\begin{itemize}
  \item mIoU:~We first employ the Mean Intersection over Union~{(mIoU)} metric to evaluate the performance on the VOC2012 dataset for the WSSS task of natural image. We use a range of thresholds to quantify the mIoU of CAM seeds separately and record the peak one representing the performance of CAM methods.
  \item FwIoU:~Following the setting of our previous study~\cite{WSSS2022han}, we utilize the Frequency weighted Intersection over Union~{(FwIoU)} metric to evaluate the performance of CAM methods on BCSS-WSSS dataset.
  \item mPxPA:~In the experiments on OpenImages30k dataset~(natural image), we utilize the mean pixel average precision~{(mPxAP)}~\cite{choe2020cvpr, choe2022tpami} metric to quantify the performance of CAM methods for WSOL task.
\end{itemize}

\begin{figure*}[t]
	\centering
	\includegraphics[width=0.9\linewidth]{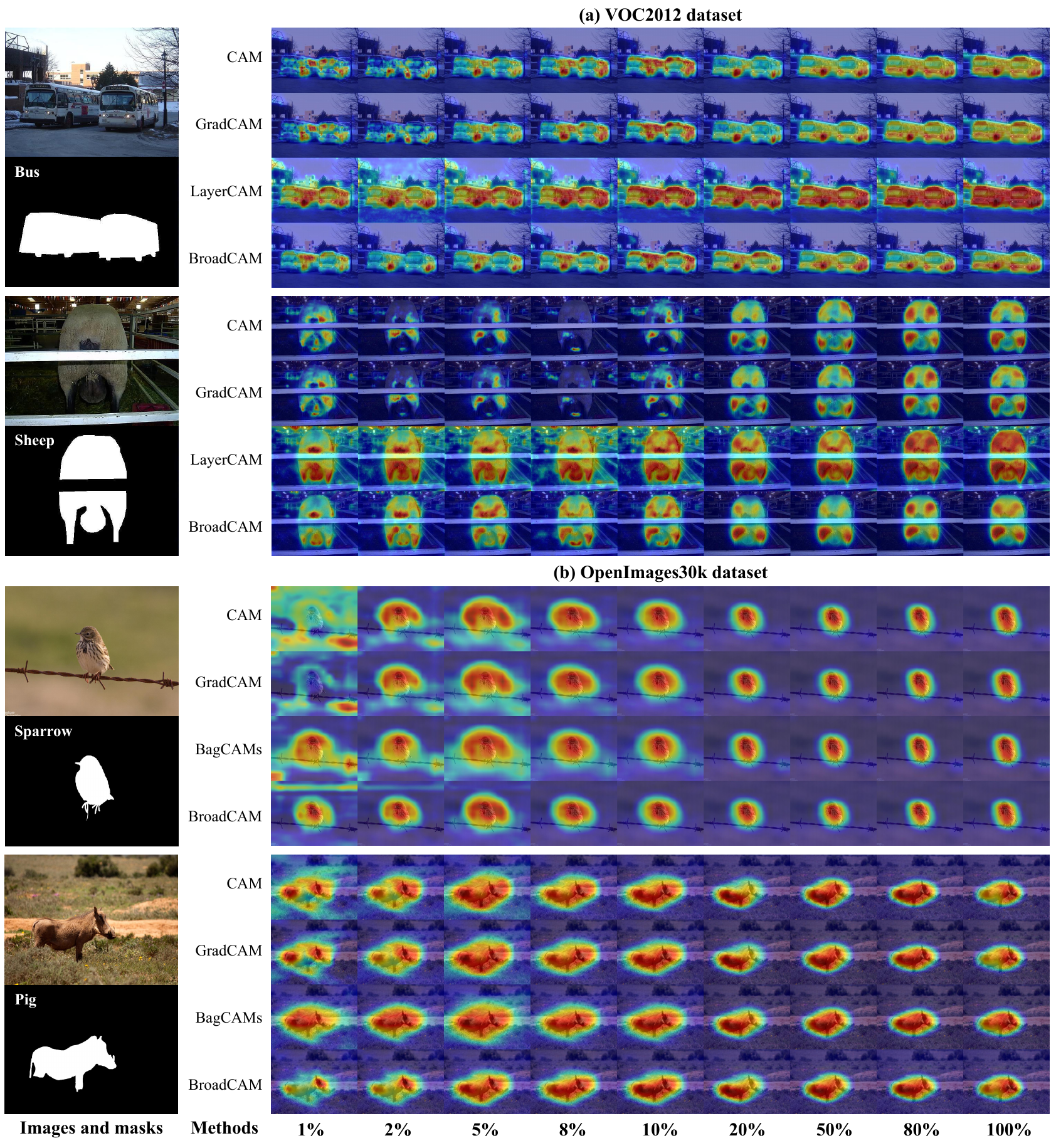}
	\caption{Comparison with CAM seeds generated by CAM methods~(CAM, GradCAM, LayerCAM, BagCAM and our {BroadCAM}) with models trained on training set of different proportions on {VOC2012 dataset} and {OpenImages30k dataset}. {Horizontally:}~$1st$ column represents the original image and mask. Then the next $9$ columns are the CAM seeds generated with models trained on training sets of different proportions~(1\%, 2\%, 5\%, 8\%, 10\%, 20\%, 50\%, 80\%, 100\%), respectively.}\label{fig:CAMs_natural}
\end{figure*}

\begin{figure}[t]
	\centering
	\includegraphics[width=1\linewidth]{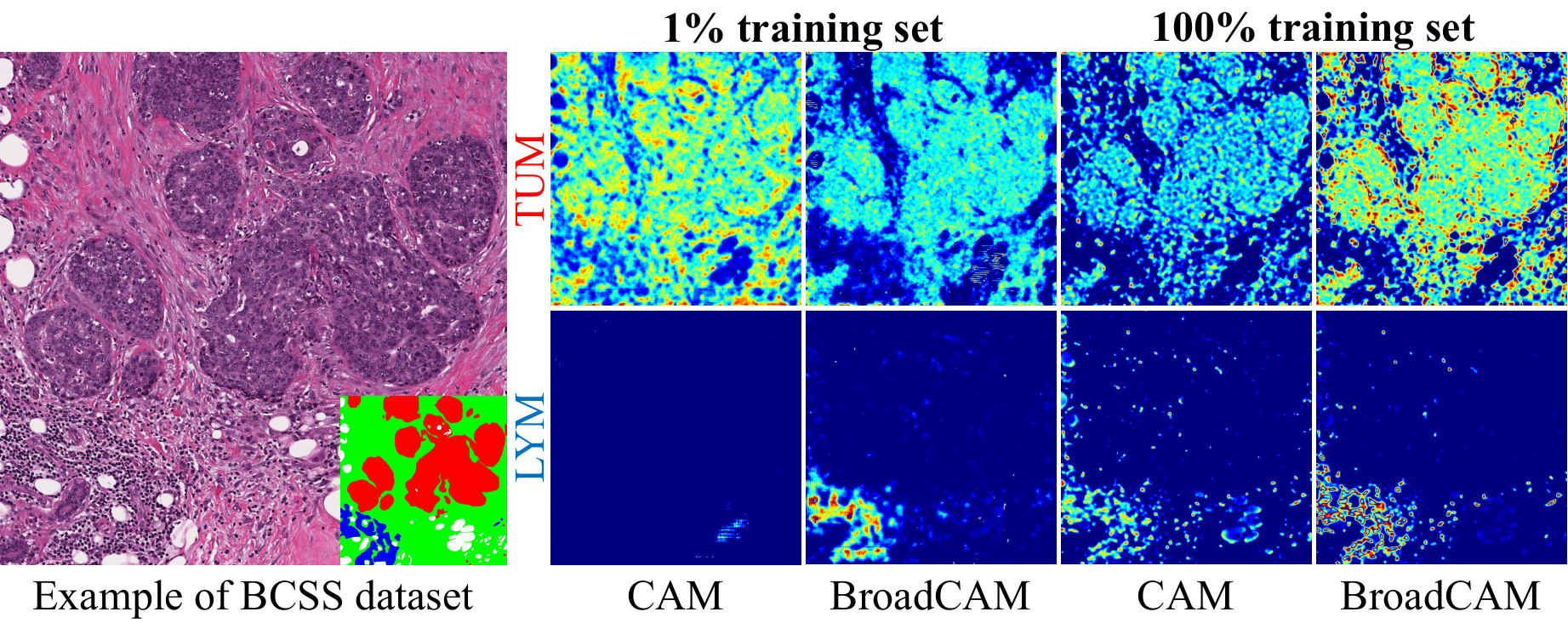}
	\caption{Visual comparison of the CAM seeds generated by CAM and our BroadCAM. An example image and its ground truth are shown on the left-hand side (Red: tumor epithelium, Blue: lymphocyte, Green: stroma). On the right side, we show the CAM seeds of two classes TUM and LYM generated by CAM and BroadCAM with 1\% and 100\% training sets.}
	\label{fig:CAMs_BCSS}
\end{figure}

\begin{table*}[t]
	\centering
	\caption{Comparison of our proposed BroadCAM and existing CAM approaches on OpenImages30k dataset with different proportions of training data. mPxAP is used for the evaluation.}
	\begin{tabular}{c|c|c|c|c|c|c|c|c|c|c|c|c}
		\hline
		\multirow{2}{*}{Dataset}& \multicolumn{1}{c|}{\multirow{2}{*}{\makecell[c]{Training strategy\\(Architecture)}}} & \multicolumn{1}{c|}{\multirow{2}{*}{Split}} & \multirow{2}{*}{CAM method} &\multicolumn{9}{c}{Performance of CAM seeds with different proportions of training samples}\\ \cline{5-13}
		
		&& &  & 1\%   &2\%   &5\%   &8\%   &10\%
		& 20\%   &50\%   &80\%   &100\% \\ \hline
		\multirow{8}{*}{\makecell{OpenImages30k\\(natural image)}}
		& \multirow{8}{*}{\makecell[c]{DA-WSOL\cite{zhu2022weakly}\\(ResNet50)}}& \multirow{4}{*}{val} & CAM
		&39.32  &50.13    &49.81    &60.10    &60.87    &62.32  &64.91    &65.41     &65.74    \\
		& &   & GradCAM
		&43.38  &53.67    &51.44    &61.92    &62.73    &63.89  &66.42    &66.98     &67.64     \\
		& &   & BagCAM
		&50.85  &57.10    &53.30    &61.92    &62.08    &62.76  &65.82    &66.33     &67.52    \\
		& &   & {BroadCAM}
		&\textbf{53.31}  &\textbf{58.75}    &\textbf{58.01}    &\textbf{63.78}    &\textbf{64.25}    &\textbf{65.18}  &\textbf{67.32}    &\textbf{67.60}     &\textbf{68.32}    \\ \cline{3-13}
		
		& & \multirow{4}{*}{test}  & CAM
		&38.47  &49.41    &49.57    &59.54    &60.35    &61.50  &64.07    &64.68     &65.05     \\
		& &   & GradCAM
		&42.63  &53.00    &51.24    &61.32    &62.00    &63.26  &66.02    &66.63     &67.12     \\
		& &   & BagCAM
		&49.54  &56.29    &52.64    &60.84    &61.38    &62.01  &65.45    &65.78     &66.89     \\
		& &   & {BroadCAM}
		&\textbf{52.71}  &\textbf{58.68}    &\textbf{57.48}    &\textbf{63.32}    &\textbf{63.86}    &\textbf{65.01}  &\textbf{67.16}    &\textbf{67.51}     &\textbf{68.10}    \\ \cline{2-13}
		\hline
	\end{tabular}\label{tab:quantitative_WSOL}
\end{table*}

\subsection{Comparisons of CAM Methods on WSSS}\label{sec:exp_WSSS}
First of all, we compare our BroadCAM with three representative CAM methods on WSSS, including the original CAM~\cite{CAM2016} and two gradient-based CAM approaches (GradCAM~\cite{GradCAM2019} and LayerCAM~\cite{LayerCAM2021}). All the experiments are conducted on a natural image dataset VOC2012~\cite{VOC2012} and a medical image dataset BCSS-WSSS~\cite{WSSS2022han}. To comprehensively investigate the performance of CAM approaches from large-scale data to small-scale data, we conduct a data gamut experiment with different proportions of the training data~(1\%, 2\%, 5\%, 8\%, 10\%, 20\%, 50\%, 80\%, 100\%). Besides the scale of the data, we also conduct comparisons on various CNN architectures and training strategies. For the experiments on the VOC2012 dataset, we employ the puzzle strategy proposed in PuzzleCAM~\cite{Puzzle2021} on ResNeSt101 and ResNeSt269~\cite{zhang2022resnest}, respectively. For the experiment on the BCSS-WSSS dataset, we apply our previously proposed training strategy, called progressive dropout attention~(PDA)~\cite{WSSS2022han}, on ResNet38~\cite{wu2019resnet38}. Since we only evaluate the quality of the CAM seeds in this experiment, we do not apply any post-processing step, like AffinityNet~\cite{ma2019affinitynet} and IRNet~\cite{ahn2019IRNet}, or second-round training by the pseudo masks. Note that, no matter how many training samples are used to train the classification model, the evaluations are performed under the entire training, validation and test sets.

Table~\ref{tab:quantitative_WSSS} demonstrates the quantitative results of different approaches to WSSS on two datasets. We can observe that CAM seeds generated by all the CAM methods with large-scale training data~(more than 50\%) exhibit exceptional performance in the VOC2012 dataset on both ResNeSt101 and ResNeSt269 architectures. Visualization in Fig.~\ref{fig:CAMs_natural}~(a) demonstrates that CAM, GradCAM and BroadCAM generate very similar CAM seeds when with large-scale data. Although LayerCAM provides high activation on the object, multi-layer design may also introduce more noisy activation. When the proportion of the training data decreases, the model training and the outcomes become less reliable. Therefore, the performance of outcome-based CAM approaches drops dramatically. Under the same experimental setting, we can observe that BroadCAM mitigates the performance degradation and constantly outperforms the outcome-based CAM approaches. For example, the performance of CAM seeds generated by the original CAM (ResNeSt269) on the validation set drops from $58.45$ to $29.95$, but BroadCAM can still achieve $49.87$. Thanks to the outcome-agnostic nature, BroadCAM will not affected by unreliable training outcomes and only focus on the relevance between the feature maps and the corresponding categories, leading to less noisy CAM seeds. As shown in Fig.~\ref{fig:CAMs_natural}~(a), outcome-based approaches show either incomplete or noisy activation. However, the CAM seeds generated by BroadCAM also show weaker activation on small-scale data than the ones on large-scale data. They are still more complete with less noise compared with the other three competitors.

To further evaluate the CAM seeds, we conduct an experiment on a medical image dataset, BCSS-WSSS. The training strategy, progressive dropout attention (PDA), applied in this experiment is proposed by our previous study~\cite{WSSS2022han} to prevent the classification model from focusing on the most discriminative areas. As shown in the lower part of Table~\ref{tab:quantitative_WSSS}, BroadCAM almost dominates the entire data gamut experiment. Although the performance of all four CAM approaches decreases when reducing the training samples. BroadCAM still greatly outperforms the other three competitors. Fig.~\ref{fig:CAMs_BCSS} demonstrates an example and the corresponding CAM seeds generated by the original CAM and BroadCAM on both 1\% and 100\% training sets. When the classification model is trained on the complete dataset, both CAM and BroadCAM can generate favorable CAM seeds. However, when the training set downscales to only 1\%, conventional CAM fails to activate lymphocytes and generates noisy tumor CAM seeds. Thanks to the outcome-agnostic design, BroadCAM is able to generate high-quality CAM seeds with more complete and less noisy activation even if the training process is unstable.

\begin{table*}[t]
	\centering
	\caption{Compared with the performance~(mPxAP) of CAM seeds generated by CAM methods on models trained using different {WSOL methods} with a training set of different proportions on {OpenImages30k dataset}.}
	\begin{tabular}{c|c|c|c|c|c|c|c|c|c|c}
		\hline
		\multirow{2}{*}{Dataset}     & \multirow{2}{*}{\makecell[c]{Proportion}}        &  \multirow{2}{*}{\makecell[c]{CAM\\methods}} & \multicolumn{2}{c|}{\makecell[c]{CAM\cite{CAM2016}}} &   \multicolumn{2}{c|}{\makecell[c]{ADL\cite{Choe2019ADL}}} &   \multicolumn{2}{c|}{\makecell[c]{Cutmix\cite{yun2019cutmix}}} &   \multicolumn{2}{c}{\makecell[c]{DA-WSOL\cite{zhu2022weakly}}}        \\ \cline{4-11}
		&          &            &   val   &   test &   val   &   test &   val   &   test &   val   &   test     \\ \hline
		\multirow{8}{*}{\makecell[c]{OpenImages30k\\(WSOL)}}
		&\multirow{4}{*}{1\%}
		&CAM          &38.05  &37.09  &32.71  &31.64  &37.35 &36.27       &39.32 &38.47  \\
		& &GradCAM    &41.27  &40.51  &35.20  &33.84  &40.66 &39.85       &43.38 &42.63  \\
		& &BagCAM         &47.47  &46.19  &38.39  &37.04  &46.74 &45.40       &50.85 &49.54  \\
		& &{BroadCAM}             &\textbf{51.88}  &\textbf{51.24}  &\textbf{48.73}  &\textbf{47.58}  &\textbf{51.23} &\textbf{50.55}       &\textbf{53.31} &\textbf{52.71}       \\ \cline{2-11}
		&\multirow{4}{*}{100\%}
		&CAM          &58.50  &57.85  &56.78  &56.37  &57.31 &56.27       &65.74 &65.05  \\
		& &GradCAM    &59.71  &59.12  &57.83  &57.36  &58.65 &57.87       &67.64 &67.12  \\
		& &BagCAM        &60.01  &59.33  &57.94  &57.48  &59.04 &58.06       &67.52 &66.89  \\
		& &{BroadCAM}             &\textbf{62.23}  &\textbf{61.68}  &\textbf{60.61}  &\textbf{60.25}  &\textbf{61.89} &\textbf{61.15}       &\textbf{68.32} &\textbf{68.10}       \\ \hline
	\end{tabular}\label{tab:training_strategy}
\end{table*}

\begin{figure*}[t]
	\centering
	\includegraphics[width=1\linewidth]{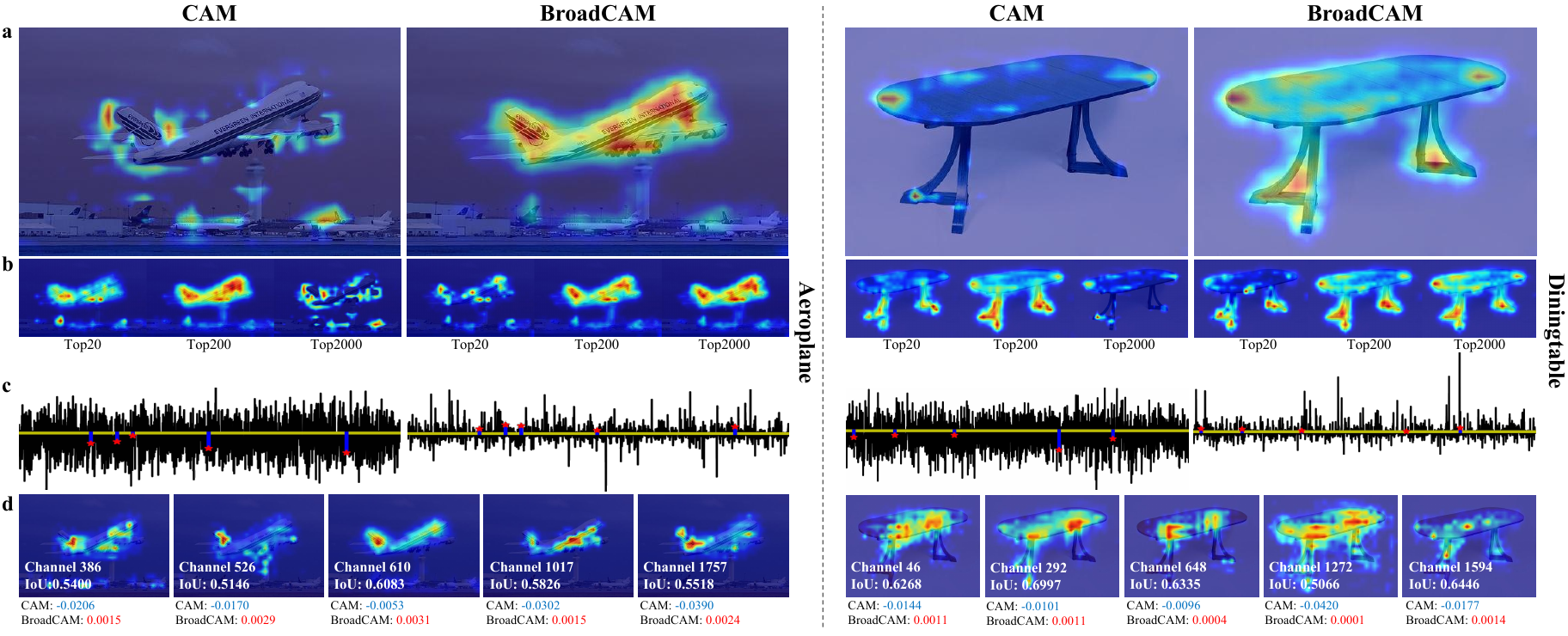}
	\caption{Visual comparison of the reliability of CAM weights. Two examples are selected from the VOC2012 dataset and the experiment is conducted on 1\% training set. (a) The CAM seeds generated by all the feature maps. (b) The CAM seeds generated by the selected feature maps of top20, top200, and top2000 weights. (c) The CAM weights in the order of the channel numbers of the feature maps, with five selected ones highlighted by red stars. (d) The selected feature maps and their corresponding channel number, IoU with ground truth and the value of the weights. Positive weights are marked in red and negative ones are marked in blue.}\label{fig:Interpretation}
\end{figure*}

\subsection{Comparisons of CAM Methods on WSOL}\label{sec:exp_WSOL}
Next, we conduct experiments to evaluate the ability of different CAM approaches on WSOL, including CAM~\cite{CAM2016}, GradCAM~\cite{GradCAM2019}, BagCAM~\cite{BagCAMs} and BroadCAM. In this experiment, we employ the training strategy DA-WSOL~\cite{zhu2022weakly} to train the ResNet50 model on different proportions of the training set in the OpenImages30k dataset. Quantitative results shown in Table~\ref{tab:quantitative_WSOL} demonstrate that the CAM seeds generated by BroadCAM are well-qualified for object localization. A similar trend in WSSS results can also be observed in WSOL ones that the performance of BroadCAM decreases more smoothly than existing CAM approaches along with the reduction of training data. Fig.~\ref{fig:CAMs_natural}~(b) visualizes CAM seeds generated by different CAM approaches with different training data scales. When with enough training data, there is no visual difference in all the CAM approaches. When reducing the training samples, existing outcome-based CAM approaches introduce false positive activation even the object can be easily separated from the background, such as the sparrow example.

To further evaluate the robustness and the flexibility of various CAM approaches on different WSOL training strategies, including original CAM~\cite{CAM2016} with no training strategy, ADL~\cite{Choe2019ADL}, cutmix~\cite{yun2019cutmix} and DA-WSOL~\cite{zhu2022weakly}. This experiment is conducted on 1\% and 100\% training sets. There is an interesting observation that not all the strategies proposed for WSOL can gain improvement compared with the original CAM baseline, which is also reported in the previous study~\cite{choe2022tpami}. Nevertheless, BroadCAM constantly outperforms existing CAM approaches in all the training strategies. And the performance gap of BroadCAM between 1\% and 100\% training sets is the least among all the four CAM methods. 
\subsection{Reliability of CAM Weights}
\label{Sec:effectiveness_analysis}
Since the CAM seeds are generated by the weighted linear aggregation of the feature maps. Therefore, whether the CAM weights are reliable will be the key to CAM generation. In this part, we conduct two visualization experiments to deeply investigate the reliability of the CAM weights, especially for small-scale data.

In the first experiment, we want to decode the relationship between the CAM weights and the feature maps. Fig.~\ref{fig:Interpretation}~(a) demonstrates two examples of CAM seeds generated by CAM and BroadCAM with a model trained on 1\% training set. We can observe that the traditional CAM fails to generate reliable CAM seeds, whereas our BroadCAM succeeds. To explain this, we first generate three CAM seeds using the selected feature maps of top20, top200 and top2000 weights. In Fig.~\ref{fig:Interpretation}~(b), we can observe that the top20 and top200 results of CAM and BroadCAM are very similar. And both CAM seeds generated by the feature maps with top200 weights can actually provide precise activation on the objects. However, in the top2000 results, we can observe an obvious activation degradation on the CAM seeds generated by the original CAM. This is because unreliable training with small-scale data will generate noisy weights for the outcome-based CAM techniques, leading to incorrect deactivation on the high class-relevance feature maps. To further justify this conclusion, we plot the CAM weights in Fig.~\ref{fig:Interpretation}~(c) and show five high class-relevance feature maps and their corresponding weights in Fig.~\ref{fig:Interpretation}~(d). Note that, the weights are in order of the channel numbers of the feature maps. We can easily observe that the original CAM deactivates the high class-relevance feature maps with negative weights, whereas BroadCAM activates them with positive weights. From this experiment, we make an assumption that the weights of BroadCAM have a higher correlation with the relevance~(IoU) of feature maps when with small-scale data, compared with the original CAM.

\begin{figure*}[t]
	\centering
	\includegraphics[width=1\linewidth]{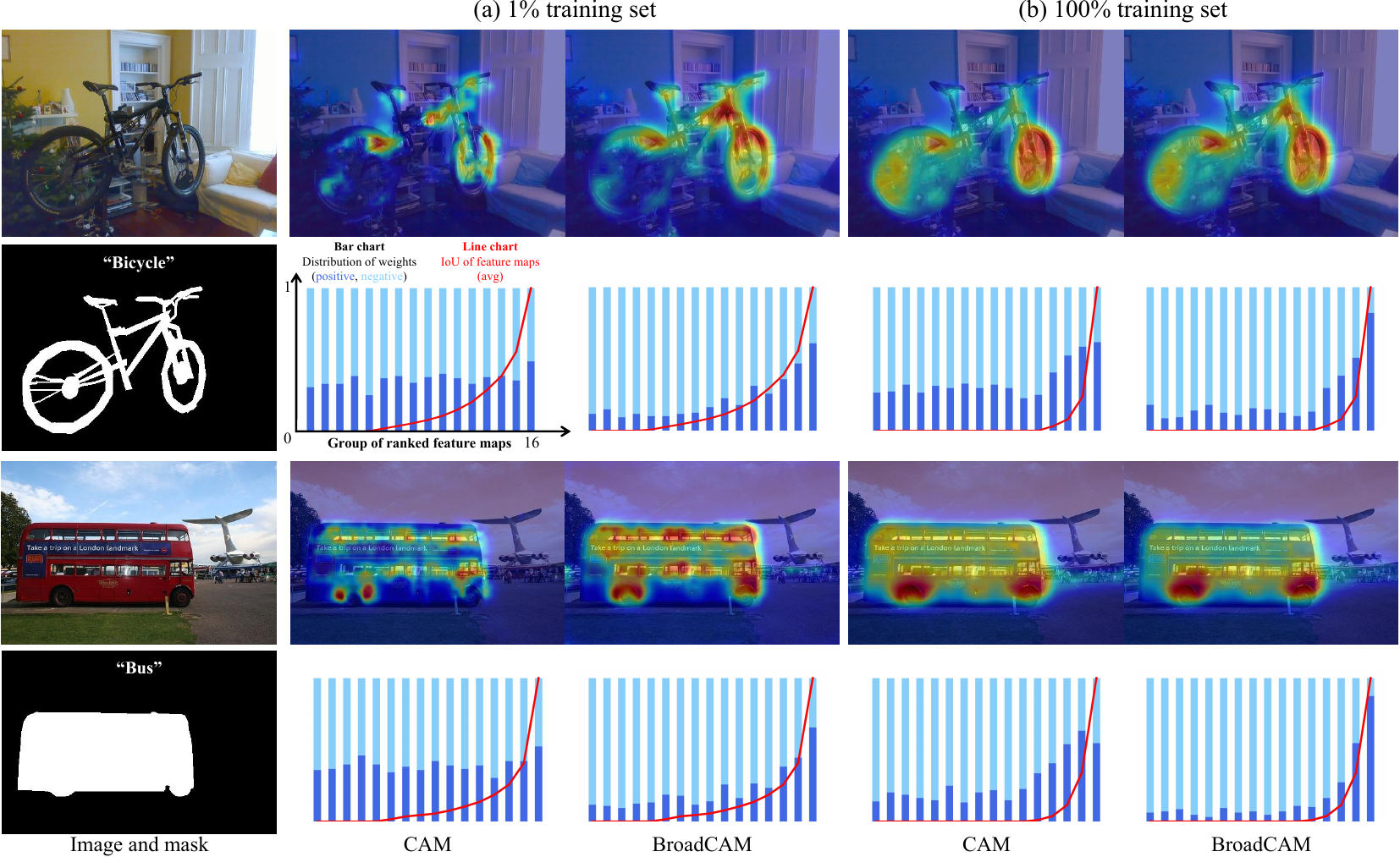}
	\caption{Visualization of the relationship between the distribution of weights and the relevance of feature maps. Two examples are selected from the VOC2012 dataset and the experiment is conducted on both 1\% and 100\% training sets. For each example, we show the original image, pixel-level ground truth, generated CAM seeds and mixed charts with line charts and bar charts. In each mixed chart, we first group the feature maps according to the rank of the IoU. The line chart (in red) shows the IoU of each group of feature maps and the bar chart (in blue) demonstrates the distribution of the weights in each group. The number of positive/negative weights is in deep/light blue.}
	\label{fig:Analysis}
\end{figure*}

To further investigate this assumption, we conduct another experiment to visualize the relationship between the distribution of CAM weights and the class-relevance feature maps. Fig.~\ref{fig:Analysis} demonstrates two examples with the generated CAM seeds on 1\% and 100\% training data. The line chart (in red) shows the IoU of the feature maps and the bar chart (in blue) demonstrates the distribution of the weights. The details of the experimental settings are demonstrated as follows. In the line chart, we first rank the feature maps by the IoU of the target class. Then we split the feature maps into 16 groups, each of which contains an equal number of feature maps. The red line is the average normalized IoU of each group of feature maps. In the bar chart, each bar indicates the total number of positive or negative weights in one group of feature maps. In each group, positive weights are in deep blue and negative weights are in light blue. By putting the bar chart and line chart together, we can easily observe that the weights generated by BroadCAM show more similar trends with the class relevance of the feature maps compared with the original CAM. For the cases in 100\% training set in Fig.~\ref{fig:Analysis}~(b), the number of positive weights is positively correlated with the class relevance for both BroadCAM and CAM, leading to favorable CAM seeds for both CAM techniques. However, when with only 1\% training data in Fig.~\ref{fig:Analysis}~(a), CAM fails to show a positive correlation but the one of BroadCAM still remains. Inaccurate activation makes CAM fail to generate clean and robust CAM seeds. This experiment somehow justifies our assumption that the weights generated by BroadCAM have a higher correlation with the relevance of feature maps compared with the original CAM, especially for small-scale data, thanks to the outcome-agnostic nature.

\subsection{Ablation Study on Multi-layer Design}\label{Sec:ablation}
Since BroadCAM is also capable of aggregating features from multiple layers. Here, we conduct an ablation study to discuss the advantages and drawbacks of the multi-layer design for BroadCAM in large-scale and small-scale weakly supervised applications. In this experiment, we train ResNeSt101 on 1\% and 100\% training sets of the VOC2012 dataset with different combinations of the layers in WSSS. L1 to L4 are from shallower layers to deeper layers. Quantitative results are shown in Table~\ref{tab:ablation}.

We first evaluate the performance of BroadCAM on the single layer. We can observe that using the feature maps extracted from the deepest layer shows the best performance as expected. Because the deeper layers deliver more semantic information and the shallow feature maps capture more textural information. When with a small-scale dataset, the feature maps are less stable and informative than with large-scale data, resulting in a significant decline in performance. Next, to evaluate the multi-layer performance, we gradually equip the other three layers on the deepest layer L4. Experimental results show that associating the last two layers achieves the best performance in both 1\% and 100\% training sets. When with large-scale data, the performance of BroadCAM on all the multi-layer configurations is similar. However, when with small-scale data, introducing shallow layers will harm the CAM generation due to the unstable feature maps. In this paper, all the experiments are conducted using the BroadCAM with the last two layers (L3$+$L4).

\begin{table}[t]
\centering
\caption{Ablation study on multi-layer of BroadCAM in WSSS with 1\% and 100\% training set on VOC2012 dataset~(mIoU).}
\begin{tabular}{c|c|c|c|c}
  \hline
   \multirow{2}{*}{Setting}        &   \multicolumn{2}{c|}{1\% training set}   &   \multicolumn{2}{c}{100\% training set}        \\ \cline{2-5}
                &   train   &   val    &   train   &   val    \\ \hline
   L1           & 17.61      & 17.75      & 21.00     & 20.65      \\ \hline
   L2           & 22.47      & 22.35      & 26.50     & 26.23      \\ \hline
   L3           & 40.42      & 40.26      & 58.03     & 56.16      \\ \hline
   L4           & 45.79      & 46.21      & 58.17     & 57.30      \\ \hline
   L3+L4        & \textbf{46.02}      & \textbf{46.40}      & \textbf{58.76}     & \textbf{57.55}      \\ \hline
   L2+L3+L4     & 45.11      & 45.57      & 58.56     & 57.36      \\ \hline
   L1+L2+L3+L4  & 44.49      & 44.76      & 58.54     & 57.35      \\ \hline
  \hline
\end{tabular}\label{tab:ablation}
\end{table} 
\section{Conclusion}\label{sec:conclusion}
In this paper, we first raise the importance of small-scale weakly supervised semantic segmentation and object localization, and propose a novel outcome-agnostic CAM method, called BroadCAM. To avoid the CAM weights being affected by unreliable outcomes due to unstable training, we introduce an independent broad learning system, which is friendly to small-scale data, to generate reliable CAM weights. Extensive quantitative and qualitative experiments on WSSS and WSOL tasks on three different datasets have proven the effectiveness and robustness of BroadCAM on both large-scale and small-scale data, whereas the existing CAM approaches are not able to achieve both. Two well-designed visualization experiments demonstrate that the outcome-agnostic design encourages BroadCAM to correctly activate more high class-relevance feature maps, resulting in less noisy and more complete CAM seeds. And the weights generated by BroadCAM are highly correlated with the relevance of the feature maps, which is the major reason why BroadCAM achieves more favorable results even when the training is unstable with small-scale data.

Since this study is the first one tailored for small-scale weakly supervised applications, and BroadCAM is the first outcome-agnostic CAM approach. We think that there exists a lot of future directions to be explored. (1) The reliability of the feature maps. In this study, we only focus on how to avoid the CAM weights being affected by unreliable training outcomes. However, unstable training will also harm the robustness of the feature maps when generating CAM seeds. Therefore, how to improve the feature representation ability for small-scale data is equally important. (2)~How to make good use of shallow layers. Although BroadCAM leverages multi-layer information to generate CAM seeds. It still mainly relies on the deeper layers with more semantic information. However, the activation of the feature maps from deeper layers is coarse and more abstract. To introduce the finer details of the object boundaries, making good use of shallow layers might be a feasible solution.

As the first baseline model on small-scale weakly supervised applications, we hope BroadCAM can bring some insights to this field and encourage more studies to discover more outstanding techniques to reduce the annotation efforts on semantic segmentation and object localization.

\ifCLASSOPTIONcompsoc
  \section*{Acknowledgments}
\else
  \section*{Acknowledgment}
\fi

This work was supported by
the Key-Area Research and Development Program of Guangdong Province, China (No. 2021B0101420006),
the Natural Science Foundation for Distinguished Young Scholars of Guangdong Province (No. 2023B1515020043),
the National Science Foundation for Young Scientists of China (No. 62102103),
the National Natural Science Foundation of China (No. 82372044, 82071892, 92267203 and 82271941),
Regional Innovation and Development Joint Fund of National Natural Science Foundation of China (No.U22A20345),
High-level Hospital Construction Project (No.DFJHBF202105),
Guangdong Provincial Key Laboratory of Artificial Intelligence in Medical Image Analysis and Application (No. 2022B1212010011).
the Science and Technology Major Project of Guangzhou (No. 202007030006),
the Program for Guangdong Introducing Innovative and Entrepreneurial Teams (No. 2019ZT08X214),
Key-Area Research and Development Program of Guangzhou City (202206030007).
The source code of BroadCAM is available at \url{https://github.com/linjiatai/BroadCAM}.

\ifCLASSOPTIONcaptionsoff
  \newpage
\fi

\normalem
\bibliographystyle{IEEEtran}
\bibliography{egbib}
\end{document}